\documentclass[twocolumn]{bmcart}% uncomment this for twocolumn layout and comment line below

\makeatletter
\renewenvironment{figure}[1][]{%
  \@float{figure}[#1]%
}{
  \end@float
}
\makeatother

\makeatletter
\renewenvironment{figure*}[1][]{%
  \@dblfloat{figure}[#1]%
}{
  \end@dblfloat
}
\makeatother

\usepackage{graphicx}
\usepackage{amsthm,amsmath}
\usepackage[utf8]{inputenc} %unicode support

\usepackage{siunitx} % own packages 
\usepackage[T1]{fontenc}
\usepackage[official]{eurosym}

\usepackage{url}
\usepackage{xcolor}
\usepackage{lmodern}

\usepackage{placeins}

\usepackage{cite}
 
\startlocaldefs
\endlocaldefs

\begin{document}

%%% Start of article front matter
\begin{frontmatter}

\begin{fmbox}

\title{Enhancing Robustness of Asynchronous EEG-Based Movement Prediction using Classifier Ensembles}

%first author 
\author[
  addressref={aff1, aff2},
  corref={aff1},% id's of addresses, e.g. {aff1,aff2}
                     % id of corresponding address, if any
% noteref={n1},                        % id's of article notes, if any
  email={niklas.kueper@dfki.de}   % email address
]{\inits{N.K.}\fnm{Niklas} \snm{Kueper}}% first author 
\author[
  addressref={aff1, aff2},
  email={kartik.chari@dfki.de}
]{\inits{K.C.}\fnm{Chari} \snm{Kartik}}% second author 
\author[
  addressref={aff2, aff1},                   % id's of addresses, e.g. {aff1,aff2}  
  email={elsa.kirchner@uni-due.de; elsa.kirchner@dfki.de} 
]{\inits{E.A.K.}\fnm{Elsa Andrea} \snm{Kirchner}}\newline % third author (corresponding author)  

%%%%%%%%%%%%%%%%%%%%%%%%%%%%%%%%%%%%%%%%%%%%%%
%%                                          %%
%% Enter the authors' addresses here        %%
%%                                          %%
%% Repeat \address commands as much as      %%
%% required.                                %%
%%                                          %%
%%%%%%%%%%%%%%%%%%%%%%%%%%%%%%%%%%%%%%%%%%%%%%

% DFKI 
\address[id=aff1]{%                           % unique id
  \orgdiv{Robotics Innovation Center},             % department, if any
  \orgname{German Research Center for Artificial Intelligence (DFKI)},          % university, etc
  \city{Bremen},                              % city
  \cny{Germany}                                    % country
} % UDE 
\address[id=aff2]{%
  \orgdiv{Institute of Medical Technology Systems},
  \orgname{University of Duisburg-Essen},
  %\street{},
  %\postcode{}
  \city{Duisburg},
  \cny{Germany}
}

% max 350 words

\begin{abstract}
\textit{Objective:} Stroke is one of the leading causes of disabilities affecting the sensory and musculoskeletal system. One promising approach is to extend the rehabilitation with self-initiated robot-assisted movement therapy. To enable this, it is required to detect the patient's intention to move to trigger the assistance of a robotic device. This intention to move can be detected from human surface electroencephalography (EEG) signals; however, it is particularly challenging to decode when classifications are performed online and asynchronously. In this work, the effectiveness of classifier ensembles and a sliding-window postprocessing technique was investigated to enhance the robustness of such asynchronous classification. \\
\noindent
\textit{Approach:} To investigate the effectiveness of classifier ensembles and a sliding-window postprocessing, two EEG datasets with 14 healthy subjects who performed self-initiated arm movements were analyzed. Offline and pseudo-online evaluations were conducted to compare ensemble combinations of the support vector machine (SVM), multilayer perceptron (MLP), and EEGNet classification models. \\ 
\noindent
\textit{Main results:} The results of the pseudo-online evaluation show that the two model ensembles significantly outperformed the best single model for the optimal number of postprocessing windows, as indicated by the number [EEGNet3 vs. SVM-EEGNet2, $p<0.01$; EEGNet3 vs. MLP-EEGNet2, $p<0.05$]. In particular, for single models, an increased number of postprocessing windows significantly improved classification performances. 
Interestingly, we found no significant improvements between performances of the best single model and classifier ensembles in the offline evaluation.  \\
\noindent
\textit{Significance:} We demonstrated that classifier ensembles and appropriate postprocessing methods effectively enhance the asynchronous detection of movement intentions from EEG signals. In particular, the classifier ensemble approach yields greater improvements in online classification than in offline classification, and reduces false detections, i.e., early false positives. As a result, our approach promises an improved applicability for the asynchronous detection of EEG-based movement intentions in realistic out-of-the-lab applications.

\end{abstract}

%%%%%%%%%%%%%%%%%%%%%%%%%%%%%%%%%%%%%%%%%%%%%%
%%                                          %%
%% The keywords begin here                  %%
%%                                          %%
%% Put each keyword in separate \kwd{}.     %%
%%                                          %%
%%%%%%%%%%%%%%%%%%%%%%%%%%%%%%%%%%%%%%%%%%%%%%

% Three to ten keywords 

%\keywords{EEG, ensemble, movement prediction, asynchronous classification, stroke rehabilitation}

% MSC classifications codes, if any
%\begin{keyword}[class=AMS]
%\kwd[Primary ]{}
%\kwd{}
%\kwd[; secondary ]{}
%\end{keyword}

%\end{abstractbox}
%
\end{fmbox}% uncomment this for two column layout

\end{frontmatter}

%%%%%%%%%%%%%%%%%%%%%%%%%%%%%%%%%%%%%%%%%%%%%%%%
%%                                            %%
%% The Main Body begins here                  %%
%%                                            %%
%% Please refer to the instructions for       %%
%% authors on:                                %%
%% https://www.biomedcentral.com/getpublished %%
%% and include the section headings           %%
%% accordingly for your article type.         %%
%%                                            %%
%% See the Results and Discussion section     %%
%% for details on how to create sub-sections  %%
%%                                            %%
%% use \cite{...} to cite references          %%
%%  \cite{koon} and                           %%
%%  \cite{oreg,khar,zvai,xjon,schn,pond}      %%
%%                                            %%
%%%%%%%%%%%%%%%%%%%%%%%%%%%%%%%%%%%%%%%%%%%%%%%%

%%%%%%%%%%%%%%%%%%%%%%%%% start of article main body
% <put your article body there

%%%%%%%%%%%%%%%%
%% Background %%
%%
\section*{Introduction}
To date, 94 million people worldwide suffer from the severe effects of stroke, and the estimated global cost of stroke reached over $890$ billion US dollars per year \cite{feigin_world_2025}. Among stroke survivors, $38\%$ suffer from disabilities which affect the sensory and musculoskeletal system \cite{ju_causes_2022}. This limits the mobility and movement ability of patients in their everyday lives. Hence, there is an urgent need for more effective and efficient post-stroke rehabilitation possibilities. To bridge this gap, traditional physiotherapy can be combined with robot-assisted stroke therapy for improving rehabilitation options ~\cite{krebs1998robot, mehrholz2018electromechanical, fazekas2019future}. In this context, active exoskeletons \cite{elsa_andrea_kirchner_towards_2022}, such as the upper-body RECUPERA Reha exoskeleton \cite{kumar2019modular}, have demonstrated their effectiveness in neuro-motor rehabilitation after stroke ~\cite{hortal2015using, noda2012brain, singh2021evidence}. \\ % Lopez-Larraz? 
Such a system can, for example, support upper-body arm movements and therefore yield the potential to enhance therapy outcomes by providing repeated movement support with proprioceptive feedback \cite{vahdat_single_2019}. % lit dazu das dies beneficial ist ? 
However, to enable such an exoskeleton-driven movement therapy, it is required to decode the patient's intention to move the disabled limb that is affected by a  stroke. There are several examples of how movement intentions can be implicitly detected through electroencephalography(EEG)-based brain-computer interfaces (BCIs) \cite{kirchner_multimodal_2014, kueper_avoidance_2024, bai_prediction_2011, sburlea_continuous_2015, liu_eeg-based_2018, kuo_classification_2011, lopez-larraz_continuous_2014, lew_detection_2012, lana_detection_2015, ceradini_effect_2025}, which enables triggering the support of an assistive exoskeleton during interaction with the system \cite{seeland_online_2013, sarasola-sanz_hybrid_2024}. \\
These movement intentions can be detected from human surface EEG-signals by decoding neural correlates such as the movement-related cortical potentials (MRCPs) \cite{shibasaki_what_2006}, especially the pre-movement components such as the lateralized readiness potential (LRP) \cite{gratton_pre-_1988, de_jong_use_1988}, i.e. the late readiness potential \cite{kornhuber_hirnpotentialanderungen_1965}, as well as the event-related desynchronization/synchronization (ERD/ERS) \cite{pfurtscheller_event-related_1999}. Many researchers have demonstrated the feasibility of detecting movement intentions by training classifiers based on these neural features (MRCPs, ERD/ERS) in healthy individuals e.g. \cite{seeland_spatio-temporal_2015, kueper_avoidance_2024, bai_prediction_2011, lopez-larraz_continuous_2014, lew_detection_2012, shakeel_review_2015} as well as in patients suffering from stroke such as in \cite{lopez-larraz_uncovering_2025, ang_clinical_2009} (for review see \cite{lopez-larraz_brain-machine_2018, yang_eeg_2022}). Mostly, after the extraction of MRCP or ERD/ERS features, traditional machine learning algorithms such as support vector machines (SVMs) or linear discriminant analysis (LDA) classifiers are trained to detect movement intentions (for review see \cite{lotte_review_2007, lotte_review_2018}). Besides these approaches, both types of features were compared \cite{seeland_spatio-temporal_2015} and also combined to enhance the performance of the classification outcome \cite{liu_hybrid_2024, luo_motor_2020}. In addition to the application of these more traditional machine learning classifiers, convolutional neural network (CNN) approaches such as the EEGNet \cite{lawhern_eegnet_2018} and the many recent variants (e.g. EEGNex \cite{chen_toward_2024} or SincEEGNet \cite{arpaia_sinc-eegnet_2023}) or the deep ConvNet \cite{schirrmeister_deep_2017} together with the current trend of using transformer-based architectures, as e.g. described in \cite{zeynali_classification_2023}, were also applied to classify EEG-signals without the need for manual feature extraction. \\
However, apart from the choice of a specific classification method, one major challenge is to detect movement intentions of attempted movements, online and in an asynchronous fashion. This means the classifier needs to continuously predict the person's intention to move, in contrast, to the classification of single or a few multiple EEG windows that are segmented based on a cue. Such a cue usually indicates the start of a motor imagination phase, as for example in the GRAZ BCI \cite{pfurtscheller_graz-bci_2003}. Despite this, our approach is to enable the detection of fully self-initiated movement attempts \cite{folgheraiter_measuring_2012, seeland_adaptive_2017} supported by an exoskeleton to provide a natural and intuitive interaction between the human and the robotic device. For this purpose, a robust and performable asynchronous online detection of movement intentions is required. However, one problem with such an approach is the robustness of the classification against false positive \cite{song_eeg-based_2022, bai_prediction_2011, kim2023asynchronous} detections, which in our robot-assisted application results in early movement detections and could lead to undesired behavior of the assistive robotic device. Concretely, this means the assistive robot (i.e., an exoskeleton) would be triggered to support movement attempts based on a falsely detected movement intention that contradicts the person's actual intention to move. This could impact the trustworthiness of patients regarding the assistive robot in future rehabilitation sessions and must be avoided at all costs. \\ 
One approach to increase the robustness of EEG-based movement intention detection is to use multiple classifiers in a classifier ensemble, which was described as early as 2007 in the field of EEG-based BCIs \cite{sun_experimental_2007} and was discussed in subsequent years  \cite{soria-frisch_critical_2013}. To date, there are multiple examples of classifier ensembles that were successfully applied to EEG data and showed promising results for a variety of classification tasks \cite{alsuradi_ensemble_2022, dhara_fuzzy_2024, alsuradi_ensemble_2023, koley_ensemble_2012, zeynali_classification_2023, abbasi_eeg-based_2021, prabhakar_ensemble_2024, hosseini_random_2018}. Specifically, for the decoding of motor intentions within synchronous motor imagery paradigms, classifier ensemble approaches have proven to be advantageous compared to single classifiers \cite{zheng_ensemble_2022, luo_motor_2020, bhattacharyya_performance_2014, dolzhikova_subject_independent_2022}.\\
However, to the best of our knowledge, the effectiveness of classifier ensembles for the asynchronous (pseudo) online detection of movement intentions in executed or attempted movements has not been investigated so far. This effect was evaluated on datasets with healthy subjects in this work. In addition, we also investigated the role of considering multiple classified EEG windows for final decision making, which has been demonstrated to be superior for motor imagery classification \cite{gaur_sliding_2021, saideepthi_sliding_2023}. However, in this work,  we analyzed this technique for postprocessing of classifications scores in asynchronous movement intention detection. \\

The contributions of this work are as follows.
\begin{itemize}
    \item We enhanced the robustness of the asynchronous detection of movement intentions by combining multiple neural network models in a classifier ensemble. 
    \item We show that the postprocessing of multiple classified EEG windows reduces early movement detections in asynchronous classifications (i.e. false positives). 
   \item We showcase the potential of both these approaches to improve post-stroke rehabilitation and the challenges of online asynchronous EEG classifications compared to offline classifications.
\end{itemize}
\medskip
The remainder of this paper is organized as follows. In the \textit{Methods} section, the analyzed EEG datasets, data processing, as well as the applied classification and evaluation methods are described together with the statistical analysis of the results. In the \textit{Results} section, the classification outcomes of the carried out offline and pseudo-online evaluations are presented and subsequently discussed in the \textit{Discussion} section. Finally, we conclude the findings and provide an outlook to future work in section \textit{Conclusion and Outlook}.

%%%%%%%%%%%%%%%%
%% Methods %% 
%%
\section*{Methods}
% start with methods 
In this section the used datasets as well as the EEG data processing and classification together with the models used for classification are described. Furthermore, the classifier ensemble, multi window postprocessing method and the principles of the carried out offline and pseudo-online evaluation are presented. Finally, the applied statistical analysis is depicted. 

\subsection*{Datasets and experimental setup}
% reference scientific reports paper and give outline - image from other paper? 
In this work, two recorded datasets, referred to as A1 and A2, which include EEG data of unilateral movement executions of a reaching task, were used for the data analysis and evaluations. Both datasets were joined together since they share the same experimental setup, procedure, and recording devices. \\
In the following sections \ref{sec:A1} and \ref{sec:A2} the datasets are described. For a  detailed description of the dataset and the experimental setup, please refer to \cite{kueper_avoidance_2024}. The described datasets are publicly available via the following  Zenodo repository: \url{https://doi.org/10.5281/zenodo.17940098}. 

\subsubsection*{Dataset A1}
\label{sec:A1}
The first dataset (A1) consisted of eight healthy participants (mean age: $25.5 \pm 4.0$) who completed an executed-movement paradigm involving a unilateral reaching task. All subjects gave their informed written consent to participate in the study, and ethical approval was given by the University of Bielefeld. The subjects were sitting in a comfortable chair, repeatedly reached for a button and pressed it with their right hand that was placed in front of them. The movement onsets that were tracked by a hand switch, as well as all other events such as pushing the switch or pressing the button etc. were recorded as events in the EEG data. The movements were completely self-paced and self-initiated by the subjects. However, there was a restriction that subjects were required to remain in a resting position for at least 5 seconds before initiating the next movement. If this restriction has not been met, an error sign consisting of a red blinking screen for a duration of 200 \si{\milli\second} was shown on a monitor that otherwise consistently showed a fixation cross with a green background. Trials including an error sign were rejected from the analysis. \\ 
Each subject performed a total of 120 movement trials, where each trial consisted of the movement period followed by the resting period as previously described. The overall movement trials were divided into three measurement sets (also called measurement runs), each of which included 40 movement trials. After each measurement set, there was a brief break to prevent subject fatigue. \\ 
EEG data was recorded from 64 active electrodes (Acticap montage) according to the extended 10-20 system and at a sampling rate of 500 \si{\hertz} using the LiveAmp64 amplifier from Brain Products. The impedances of each electrode were kept below a threshold of 5 \si{\kilo\ohm}. The data was pre-filtered by the measurement hardware in a frequency range from 0.1 to 131.0 \si{\hertz}. Additionally, EMG data was recorded with 16 EMG electrodes from pico EMG sensors from Cometa as well as motion tracking data using a Qualisys motion tracking system, for in-depth data analysis. 

\subsubsection*{Dataset A2}
\label{sec:A2}
The second dataset A2 was recorded in a consecutive study after dataset A1 was produced. In this study, six healthy subjects (mean age: 23.8 $\pm$ 0.75 years) performed the same movement task with the same experimental setup, protocols, and recording hardware as for dataset A1. However, it should be mentioned that the study was conducted in a normal lab environment, whereas the study of dataset A1 was recorded in a shielded cabin as described in \cite{kueper_avoidance_2024}. This does not affect the data analysis and evaluation, since the data in this work is processed subject-wise in an intra-subject evaluation design as described below. \\
All subjects gave their written informed consent to participate in the study, and the study was approved by the ethical committee of the University of Duisburg-Essen. 
%data availability !! 

\subsection*{Preprocessing and Feature Extraction}
In a first step, 16 out of the 64 EEG channels were selected based on a previous evaluation (see \cite{kueper_avoidance_2024}) and considering a trade-off between preparation times in the envisioned stroke rehabilitation application and classification performance. The selected channels were FZ, CZ, CPZ, PZ, P1, CP1, C1, FC1, F1, F3, FC3, C3, CP3, CP5, C5, FC5. 
After the channel selection, the EEG data was epoched trial-wise in a time range from -5.0 \si{\second} to 0.2 \si{\second} with respect to the movement onset at 0 \si{\second}. This was done to further process the complete movement trial for the different types of evaluations. It should be noted that this time range was selected to capture the complete trial from the whole resting period over the movement planning phase to the movement execution. In the next step, overlapping windows with a length of 1 \si{\second} were cut out every 0.05 \si{\second} for the separated test data and every 0.02 \si{\second} for the training and validation data. This was done to increase the number of windows available for training a model or classifier, while ensuring sufficient time for processing each window in an online classification, simulated by the test set here. The same windows were cut for each of the three classification methods described below. After windowing of the EEG data, each window was processed independently.

\subsubsection*{SVM and MLP Processing:} 
For the classification pipelines with an SVM and an MLP, a second-order Butterworth bandpass filter (0.3 to 5.0 \si{\hertz}) was applied to extract the low-frequency MRCP components in the time domain, such as the LRP and the motor potential (MP). Furthermore, the unprocessed EEG windows were also stored for feature extraction in the frequency domain. Therefore, both the time domain features of the elicited MRCPs \cite{shibasaki_what_2006} as well as oscillatory features in the frequency domain, mainly based on the ERD/ERS \cite{pfurtscheller1999event}, were considered and combined with time domain features. The further processing and feature extraction of the EEG windows were similar between the SVM and MLP pipeline, except for an xDAWN spatial filter \cite{rivet_xdawn_2009} that was additionally applied for dimensionality reduction in the SVM pipeline. \\
For the extraction of time domain features, 7 equally spaced samples from every 50 \si{\milli\second} of the last 300 \si{\milli\second} of each EEG window were extracted from the 16 remaining EEG channels that were selected. \\
For the frequency-domain features, power spectral density (PSD) was computed for all remaining EEG channels from the last 500 \si{\milli\second} of each EEG window. The PSD values were computed for each EEG frequency band (0.5-4; 4-8; 8-13; 13-30; 30-100 \si{\hertz})  using the multitaper method. Therefore, five PSD values were obtained for each channel, and the features were combined with the extracted time-domain features to form a single feature vector. The features were normalized afterward to have a mean of zero and a standard deviation of one before being fed into the SVM and MLP. 

\subsubsection*{EEGNet Processing:}
For the preprocessing of the windows for classification with the EEGNet model \cite{lawhern_eegnet_2018}, the windows were also filtered using a second-order Butterworth bandpass filter, but in a range of 0.3 to 40.0 \si{\hertz}. After filtering, the EEG windows were normalized channel-wise to have zero mean and a standard deviation of one before being fed into the EEGNet model. 

\subsection*{Classification}
\label{sec:Classification}
Three different machine learning models, namely the SVM, MLP, and the EEGNet, as well as a dummy classifier, were used for the classification of arm movement onsets of a reaching task by detecting movement intentions as well as the transition to movement execution. Therefore, the binary classification task was to distinguish between resting (negative class) and movement preparation/execution (positive class) to precisely detect movement onsets and trigger support of an upper-body exoskeleton for the envisioned robot-assisted stroke rehabilitation therapy. \\
All classification methods were trained on selected EEG windows, which were specified as: \\ $[-3, -2, -1.5, -1.0, -0.8, -0.6]$  \si{\second} for the negative class and $[0.04, 0.06, 0.08, 0.1, 0.12, 0.14]$ \si{\second} for the positive class. Here, each window is named after the time where it ends in relation to the movement onset (e.g., window -1.0 \si{\second} ranges from -2.0 \si{\second} to -1.0 \si{\second}). \\
Each classification method was trained using a leave-one-set-out cross-validation technique, where two measurement sets with a total of 80 trials were used for training, and the remaining set with 40 trials was split equally for validation and testing. Therefore, the training was performed on a total of 960 windows (12 windows of 80 trials). The classifiers were stored after training for further evaluation. In the following, the details of each classifier are described. 

\subsubsection*{SVM:}
The SVC implementation from the sklearn Python package was used with a linear kernel as the SVM classifier. The complexity parameter of the SVM was tuned as a hyperparameter using grid search with a 5-fold internal cross-validation, with a grid of values [1e-6, 1e-5, 1e-4, 1e-3, 1e-2, 1e-1, 1, 10]. For a reference to this pipeline, refer to \cite{kueper_avoidance_2024,kirchner_multimodal_2014}.

\subsubsection*{MLP:}
We developed an MLP neural network model used for classification. The model was designed for the classification of movement intentions as an alternative to SVM and was initially intended for classifying time-domain features by replacing xDAWN + SVM with a single model. It was then extended to classify both time and frequency-domain features. \\
The model comprises a trainable normalization layer and three densely connected layers with 32, 20, and 12 neurons and leaky ReLU activations (alpha = 0.5). The output layer contains a single neuron and sigmoid activation for the binary classification. Between the dense layers, dropout layers (dropout rate = 0.5) and batch normalization layers were added for regularization. The source code of the model's architecture is made publicly available and can be found in the following GitHub repository: \url{https://github.com/nkueper/EEG_MLP_model}. 

The model was then trained for a maximum of 200 epochs with the binary cross-entropy loss function and an Adam optimizer. The batch size was set to 16. Additionally, an early stopping technique was used, where the patient parameter was set to 70, and the best weights were restored after training. 

\subsubsection*{EEGNet:}
EEGNet is an established and widely used neural network model for EEG classification. In our analysis, the original training parameters were used as described in \cite{lawhern_eegnet_2018}.  However, after the hyperparameters of the model were optimized on our data, the kernel size of the temporal filter was set to 50. In addition to the original architecture, a normalization layer was added, similar to the MLP model. Through this, a trainable channel-wise standardization of the input data was ensured and included directly in the model architecture. 

\subsubsection*{Dummy:}
A dummy classifier was set up as a comparison to the evaluated classifiers. Here, the untrained MLP model, which was only randomly initialized, was used as the dummy classifier. The classifier was then used to make predictions on the same features as the actual MLP model. 

\subsection*{Ensemble Approach and Postprocessing}
The proposed ensemble approach combines the results of multiple classifiers to aim for more robust predictions of movement onsets. This is particularly important in online scenarios where EEG classification is performed asynchronously. Here, the assumption was that multiple classifiers, particularly with different architectures and characteristics, may not necessarily yield the same predictions for the same EEG window and thus may be more robust if they are combined. Therefore, we multiplied the output class probabilities for different combinations of the described classifiers and used the resulting probability to determine the class label for each classified EEG window. To enable a fair comparison for a different number of classifiers that are combined, the decision boundary, which is at 0.5 for a binary classification task, was adapted to $0.5^n$ where $n$ is the number of classifiers combined. \\
Although the main focus was to evaluate this approach in a online asynchronous classification scenario, we also applied this approach to an offline classification scenario for comparability reasons. In both cases, only the evaluation principle (described below) differed but not the classification methodology (feature extraction, ensemble approach etc.). \\ 
Besides the combination of multiple classifiers, the number of windows that consecutively need to be classified as the positive class label to manifest a detected movement onset was also evaluated in a pseudo-online evaluation. The number of evaluated sliding windows in a row ranged from one to three windows. Only if all predicted windows were instances of the positive class, a movement onset was detected.

\subsection*{Performance Evaluation}
To evaluate the performance of the ensemble classifier approach, an offline and a pseudo-online evaluation scheme were developed for comparing both types in this context. Here, the offline evaluation consists of the performance evaluation of single EEG windows, as it is performed mostly in BCI research, whereas in the pseudo-online evaluation, we aimed at a realistic online classification scenario for the continuous asynchronous detection of movement intentions. 
Therefore, the classification methods, which included all classifiers (Dummy(D), MLP(M), SVM(S), EEGNet(E)) and all ensemble combinations of them, which were SVM-MLP(SM), SVM-EEGNet(SE), MLP-EEGNet (ME), as well as SVM-MLP-EEGNet (SME), were compared under both evaluation schemes. 

\subsubsection*{Offline Evaluation:}
In the offline evaluation, the performance of each classification method was evaluated on the individual EEG windows from the separated test set. The windows used for this evaluation were from the same time range as those used for training the classifiers (see section \ref{sec:Classification}). Accuracy was used as a metric to evaluate the performance of the classified windows since the number of windows in each class was balanced. Using the leave-one-set-out cross-validation, 42 classification results (14 subjects x 3 sets) were obtained for comparison of the classification methods.

\subsubsection*{Pseudo-Online Evaluation:}
In the pseudo-online evaluation, the performance of the classification methods was analyzed by creating a realistic EEG processing and evaluation scheme, in which a real online application with an assistive robotic exoskeleton was simulated. Therefore, the EEG classification of movement onsets was treated as if it would directly trigger an assistive exoskeleton in real time based on the classification output of each of the compared classification methods. Furthermore, the optimal number of sliding windows that were considered for the final decision-making process was evaluated as a postprocessing parameter. Due to this, the final decision for a movement onset was made based on the specified number of EEG windows that were consecutively classified as \textit{movement intention} using this postprocessing method. The number of EEG windows evaluated was varied between one and three windows. \\
To evaluate the performance of the pseudo-online classifications under the described assumptions about the real application scenario, a custom metric was used in this evaluation. Here, three kinds of possible actions that follow the output of the final decision-making were distinguished for each movement trial, which were A) correct detection, B) early detection, and C) no detection. A correct detection was defined as the classification method's decision for the \textit{movement intention} class (after postprocessing) within the range from - 0.75 \si{\second} to 0.15 \si{\second} relative to the actual movement onset. This time range was motivated by considering the application scenario, where very early movement intention detections, although possible \cite{bai_prediction_2011}, are not desirable as they may be perceived as unintended or at least unintuitive by the user due to unexpected, early actions of the assistive device. Furthermore, any potential delay in the EEG-based detection of a movement onset should ideally be imperceptible to the user during human–robot interaction. This must be ensured so as not to distort the person’s sense of being able to move freely by their own will. Accordingly, an early detection refers to detecting a movement onset before –0.75 \si{\second}, whereas no detection indicates that no movement intention was detected during a trial. Since all trials were considered in the evaluation, we calculated a trial-wise performance (TWP), defined as the number of correct detections divided by the total number of trials evaluated. Therefore, this metric corresponds to a performance rate, ranging between 0 and 1. In addition, an early detection rate (EDR) was computed from the test trials and was used as an additional metric alongside the TWP for incorrectly classified trials. This was done to avoid early detections as much as possible, while also achieving an overall high TWP score, i.e., a high rate of correctly classified trials. \\
The types of errors and the evaluation scheme are illustrated in the following Figure \ref{evaluation_scheme}. 

% figure of evalaluation principle 
\begin{figure}[h]
 \centering
        \includegraphics[width=0.5\textwidth]{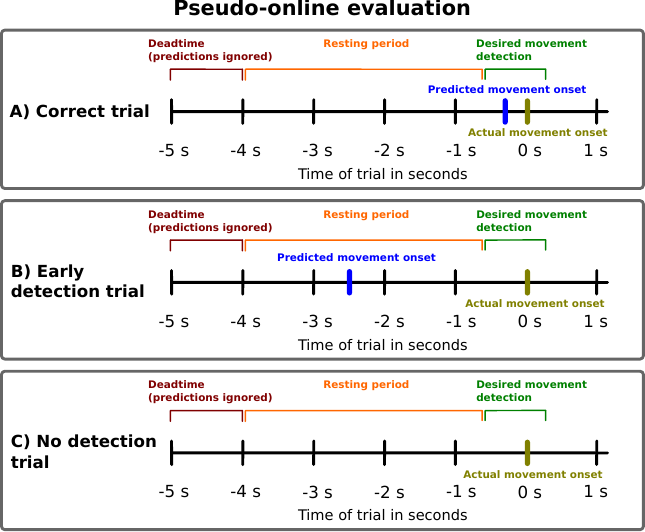}
 \caption{\textbf{Pseudo-online evaluation scheme:} It was distinguished between A) correct trials (top), B) early detection trials (middle), and C) no detection trials (bottom). In red, the deadtime period (-5s to -4s) is shown, where possible predictions are ignored by the decision logic, and in orange, the resting period is shown (-4s to -0.75s), where no movement onsets should be predicted. Finally, in green, the target period (-0.75s to 0.15s) is shown, during which a detection of movement onset is expected.}
\label{evaluation_scheme}
\end{figure}

\subsection*{Statistical Analysis}

The statistical analysis of the achieved performances was conducted using the SPSS software. The analyzed data for both the offline and pseudo-online evaluations did not entirely follow a normal distribution (tested with the Kolmogorov-Smirnov and Shapiro-Wilk tests, as well as graphical analysis). Therefore, non-parametric tests were applied in a repeated-measurements design for the inference statistics. The analyses of both evaluations are described below. 

\subsubsection*{Offline Evaluation:}
In the offline analysis of classification performance (accuracy), the within-subjects factors \textit{classification type} and \textit{number of models} were investigated to evaluate the influence of the type of classification methods and number of classifiers (single versus ensembles). This was done sequentially by first comparing the \textit{classification type} for each \textit{number of models}, which were single models and two-model ensembles, respectively. It should be noted that there was only one three-model ensemble, thus no comparisons were required for this condition. 
Therefore, a Friedman test was applied within the conditions of single models and two-model ensembles (3 levels; M-S-E for single models; SM-SE-ME for two-model ensembles). After a significant Friedman test, the Wilcoxon Signed-Rank test was used for all pairwise comparisons. Finally, a Bonferroni correction was applied to correct for multiple comparisons. \\
After the comparisons of the \textit{classification type} for each \textit{number of models}, an overall evaluation of the optimal \textit{number of models} (3 levels; single model - two model ensemble - three model ensemble) was carried out by comparing the best classification models/ensembles regarding the factor \textit{number of models}. If no single best-performing model/ensemble was found (i.e., if significant differences were found), multiple analyses of the potential best-performing models were conducted. Again, a Friedman test followed by a Wilcoxon Signed-Rank test for pairwise comparisons was applied with a Bonferroni correction.

\subsubsection*{Pseudo-Online Evaluation:}
In the pseudo-online evaluation, the TWP scores were analyzed for the same within-subjects factors \textit{classification type} and \textit{number of models} as in the offline analysis. Furthermore, the same sequential evaluation principle as in the offline evaluation was applied. However, in the pseudo-online evaluation, the additional postprocessing factor \textit{number of windows}, which was analyzed for the final decision making, was investigated before the described evaluation principle was applied. This means the optimal \textit{number of windows} were evaluated for each classification method and for each number of models individually, before the evaluation of the factors \textit{classification type} and \textit{number of models} followed.
Additionally, the EDR scores were compared only for the overall evaluation of the best models for the different \textit{number of models} to investigate the differences between single classification methods and classifier ensembles on the types of prediction errors, besides the investigation of the overall TWP scores as the main indicator of overall pseudo-online performance. 

\section*{Results}
% Here are the images of the results 
\subsection*{Offline Evaluation}
\begin{figure*}
 \centering
        \includegraphics[width=1.0\textwidth]{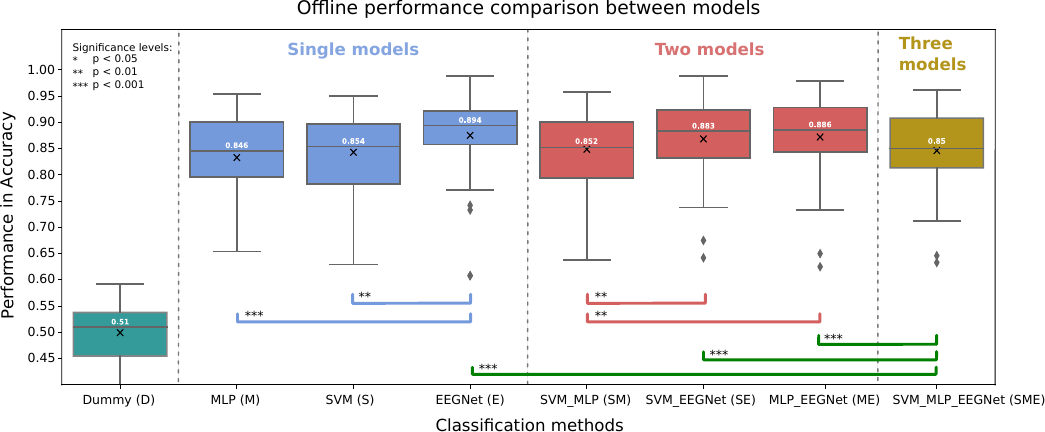}
 \caption{\textbf{Offline performance comparison of all classification methods:} Besides the dummy classifier (turquoise), the results and statistical comparison of single models (in blue), two models (in red), as well as the results of three models (in gold) are illustrated. In green, the comparisons of the best-performing models (E-SE-SME and E-ME-SME) are shown. The significance levels are shown in the top-left corner of the figure. Only significant results are indicated; if not indicated, the comparison was not significant.}
\label{offlineresults}
\end{figure*}

In Figure \ref{offlineresults}, the classification results (accuracy) of the offline evaluation are shown. 

The performances of all single models, as well as two-model and three-model ensembles, are illustrated together with the dummy classifier performing at a chance level. From the single models, the EEGNet (E) outperformed both MLP (M) and the SVM (S) $[M-E, p<0.001; S-E, p<0.01]$ with a high overall median accuracy of 0.894. For the two-classifier ensembles, there were no significant differences found between the SE and ME ensembles. However, both SE and ME significantly outperformed the SM ensemble $[SM-SE, p<0.01; SM-ME, p<0.01]$, both with manual feature extractions, as compared to the EEGNet as a CNN-based feature extraction model. 

The results of the analysis of the factor \textit{number of models} are also shown in Figure \ref{offlineresults}. Here, the highest performing single model E as well as the two model ensembles SE and ME significantly outperformed the three model ensemble $[E-SME, p<0.001; SE-SME, p<0.001; ME-SME, p< 0.001]$ marked in green. However, there were no differences between the best single model E and the two model ensembles SE and ME.

\subsection*{Pseudo-Online Evaluation}
\begin{figure*}
 \centering
        \includegraphics[width=1.0\textwidth]{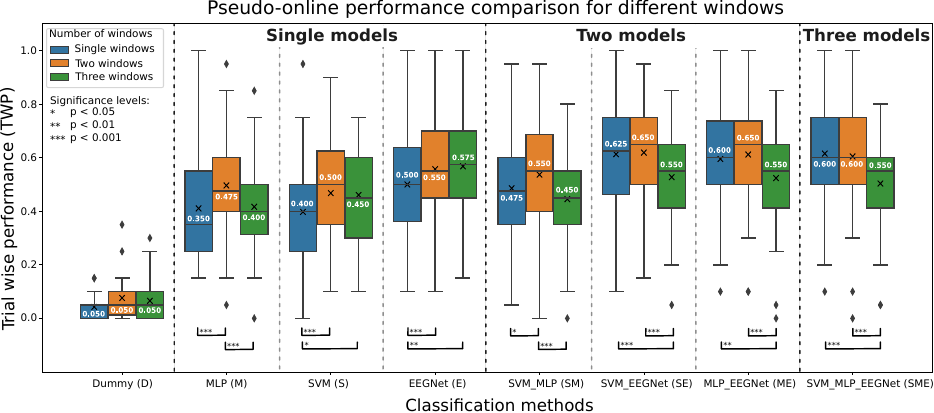}
 \caption{\textbf{Pseudo-online performance comparison for different numbers of postprocessing windows:} Besides the dummy classifier, the results of all single models, two model ensembles, and three model ensembles are shown. The coloring indicates the number of postprocessing windows used. The results of the statistical analysis of the model-wise number of postprocessing windows are indicated with black U-shaped brackets. The significance levels are located on the top left of the figure. Only significant results are indicated, if not indicated, the comparison was not significant.}
\label{overall_pseudo_online}
\end{figure*}

The overall results of the pseudo-online evaluation are shown in Figure \ref{overall_pseudo_online}.

Here, the number following the model combinations specifies the number of windows used for the postprocessing (e.g. SM2 relates to the model SVM-MLP with 2 windows used for postprocessing). 
The comparisons of the factor \textit{number of windows} show that for single models, a number of two windows used, significantly outperforms the others $[M1 - M2, p<0.001; M2 - M3, p<0.001]$ or at least outperforms a single window $[S1-S2, p<0.001; E1-E2, p<0.001]$ even though there are no significant differences to three windows for S and E.  However, descriptively, the number of three windows yields the highest median performance for E, and the number of two windows for S (see Figure \ref{overall_pseudo_online}).
For the two model ensembles, a number of two windows significantly outperforms the others for SM $[SM1 - SM2, p<0.05; SM2 - SM3, p<0.001]$ or at least outperforms the three windows for SE and ME $[SE2-SE3, p<0.001; ME2-ME3, p<0.001]$ even though there are no significant differences to one window for SE and ME. Still, the number of two windows yielded the highest median accuracy for both SE and ME. 
For the three-model ensemble (SME), the use of two windows significantly outperformed the use of three windows $[SME2-SME3, p<0.05]$ but not one window. Here, there were no differences in the median accuracy observed. \\
Conclusively, the model variants M2, S2, E3 for single models as well as SM2, SE2, and ME2 for two models, and SME2 for three models were considered as the highest performing regarding the further results. 

\begin{figure*}
 \centering
        \includegraphics[width=1.0\textwidth]{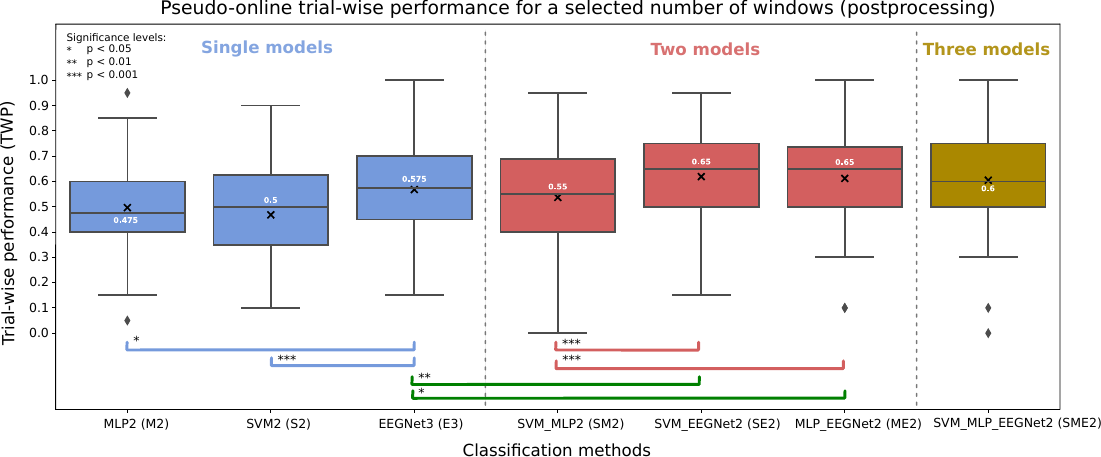}
 \caption{\textbf{Pseudo-online performances for the selected number of postprocessing windows:} The results and the comparisons of the statistical analysis of single models (in blue), two models (in red), as well as the results of three models (in gold) are illustrated. In green, the comparisons of the best-performing models (comparisons E3-SE2-SME2 and E3-ME2-SME2) are shown. The significance levels are located on the top left of the figure. Only significant results are indicated, if not indicated, the comparison was not significant.}
\label{fig4}
\end{figure*}

The classification results of all best classification methods for the conditions single models, two models, three models, and for the selected number of postprocessing windows are shown in Figure \ref{fig4}. 

The results of the analysis of the factor \textit{classification methods} show for the single models that there are significant differences between M2 and E3 $(p<0.05)$, as well as S2 and E3 $(p<0.001)$. This indicates the advantages of the EEGNet (E3) over the SVM (S2) and the MLP(M2) in the pseudo-online classifications. 
For the two model ensembles, we found significant differences between SM2 and ME2 $(p<0.001)$ as well as SM2 and SE2 $(p<0.001)$, which indicates improvements of S and M when combined with E. 
Regarding the analysis of the factor \textit{number of models}, the comparison of the best performing models E3, ME2/SE2 and SME2 showed that there are significant differences in TWP between E3 and SE2 $(p<0.01)$, as well as E3 and ME2 $(p<0.05)$. This indicates that the ensemble combination of the SVM(S) or the MLP(M) with the EEGNet(E) yields significant improvements, whereas the three-model ensemble did not outperform the single models. 

\begin{figure*}
        \includegraphics[width=1.0\textwidth]{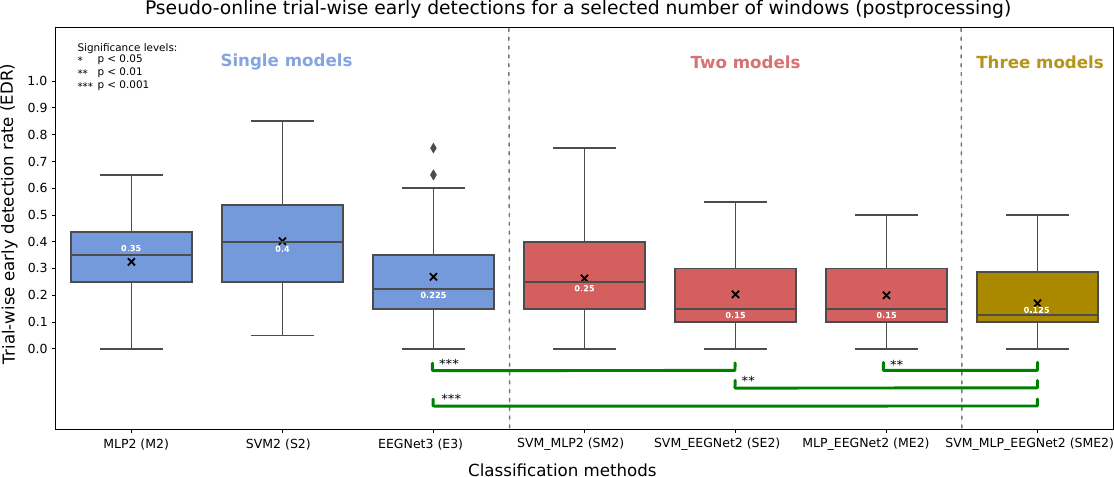}
 \caption{\textbf{Pseudo-online early detections for all classification methods:} The results and the comparisons of the statistical analysis of single models (in blue), two models (in red), as well as the results of three models (in gold) are illustrated. In green, the comparisons of the best-performing models (comparisons E3-SE2-SME2 and E3-ME2-SME2) are shown. The significance levels are located on the top left of the figure. Only significant results are indicated, if not indicated, the comparison was not significant.}
\label{fig5}
\end{figure*}

The results of the analysis of the EDR of all best classification methods from the condition single models, two models, three models, and for the selected optimal number of postprocessing windows are shown in Figure \ref{fig5}.

The results clearly show, that the two- and three-model ensemble approaches reduce the EDR scores (lower is better) in all cases when compared to a single model (see Figure \ref{fig5}). We found significant differences between all comparisons made $[E3-SE2, p<0.001; SE2-SME2, p<0.01; E3-SME2, p<0.001, E3-ME2, p<0.001; ME2-SME2, p<0.01]$. This means we found significantly lower EDR scores for the two model ensembles compared to the single model ensemble, as well as between the three model ensembles compared to the two-model ensemble(s) and the single model. 

\FloatBarrier

\section*{Discussion}
% general summary + general findings 
In this work, we investigated a combination of EEG classifier ensembles and a multi-window postprocessing method to enhance the robustness of asynchronous (pseudo) online movement predictions from the EEG. These methods were analyzed by taking into account that such an approach could be applied for improving robot-assisted stroke rehabilitation, where movements of the robotic device are triggered by detected movement intentions from the EEG. 
When interpreting the results of this work, one general finding is the major gap in classification performance between the offline and pseudo-online evaluation. Here, the median performances of the single models, which were evaluated window-wise in the offline case, were all above 0.84 (accuracy). However, when the same models were applied in the pseudo-online evaluation, realistic trial-wise performances started at 0.35 (TWP) and reached a maximum of 0.575 (TWP). This demonstrates the high difficulty of this asynchronous online classification task and the challenges for an out-of-the-lab use of such approaches. 

% Interpretations and implications 
The results of the offline evaluation show that EEGNet outperformed both the MLP and the SVM under the single model condition (see Figure \ref{offlineresults}). This may be due to the advantages of the architecture and the deep learning-based spatio-temporal feature extraction as compared to the classical feature extractions used with the MLP and the SVM. 
%This is unsurprisingly, since many authors have reported an increase in performance when comparing the EEGNet model to more classical machine learning methods such as the SVM in an offline fashion 
Interestingly, the single EEGNet model achieved the highest median accuracy in the offline evaluation and even significantly outperformed the three model ensemble (SME). Furthermore, the single EEGNet model did not perform significantly worse than the two model ensembles (ME and SE). This indicates that an ensemble approach might not be beneficial for an offline classification of movement intentions from EEG. This could be because one well-performing model alone is capable enough of making predictions on single EEG windows, and combinations of multiple prediction probabilities may lead to higher uncertainties than actual improvements in such a case. 

In contrast, the results of the pseudo-online evaluation show that there is a significant performance improvement when using the two model ensembles (SE2 and ME2) based on the methods with manual feature extraction (SVM and MLP) combined with the EEGNet model. The reason for this might be that a combination of both types of classification methods increases the robustness of the decision-making process in the pseudo-online case. Compared with an offline evaluation, such robustness across multiple consecutively classified EEG windows used in the final decision-making is more important than ensuring high reliability on a few classified EEG windows. This was further motivated by the significant reduction in EDR scores with an increased number of models combined for classification. This demonstrates the effectiveness of classifier ensembles to reduce early detections (i.e., false positive classifications) in online asynchronous movement prediction. This implies, that classifier ensembles can be beneficial to increase robustness in cases where multiple EEG windows are classified consecutively. \\
Furthermore, the results show that a number of two or three windows used in the postprocessing of the classification results can lead to an improved performance, depending on the evaluated number of classifiers (see Figure \ref{overall_pseudo_online}). However, we found that the model combinations ME2 and SE2, which are the two-model ensembles with two postprocessing windows, yield the highest performances. \\
In summary, the applied ensemble approach, as well as the postprocessing method with multiple windows, led to significant improvements in comparison to a single model and single window approach and thereby enhanced the robustness of the pseudo-online prediction of movement intentions. 

% limitations 
However, in this work, the EEG data was processed in a pseudo-online fashion (besides the offline case as a baseline comparison) instead of an evaluation in a complete online setting with an assistive robot. Although a similar processing pipeline was tested with two healthy subjects in such a realistic online application scenario, this needs to be evaluated systematically in the future.
% \textcolor{red}{Was hier jetzt kommt ist doch eher outlook??? sollte also ands Ende? Tield er Conclusion?} Since it was not a goal of this work to carry out an extensive multi-feature or multi-classifier comparison in terms of a benchmark, future work could address or investigate different feature and classification method combinations of ensemble approaches more comprehensively. Therefore, such an analysis could unveil the full potential of ensemble classifier approaches in asynchronous EEG classification. 

\section*{Conclusion and Outlook}

% Conclusion words
In conclusion, in this work, we investigated the use of classifier ensembles and a sliding window postprocessing method to enhance the robustness of online asynchronous EEG-based movement intention detection for future out-of-the-lab applications. Here, we considered an application scenario in which the general aim is to trigger movements of an active upper-body exoskeleton by decoding a person's movement intention from EEG. The results showed that both approaches, the ensemble classifier combination as well as the consideration of multiple predicted EEG windows in the postprocessing, led to significant improvements in performance and reduced early detections (i.e., false positive classifications). Considering the best single model in the pseudo-online classification, which performed at 0.5 TWP (E1, median), we achieved an improvement to a score of 0.65 TWP (SE2 and ME2, median) by combining both approaches. Furthermore, it was demonstrated that the improvements were due to a significant reduction in EDR scores when using classifier ensembles for final decision making. This means that the risk of critical unintended movement initiation and execution by a BCI-controlled robotic device can be reduced in future robot-assisted rehabilitation sessions, potentially increasing the reliability of such an approach. 

% outlook 
Due to the promising results, we plan to apply these approaches in a realistic robot-assisted stroke rehabilitation scenario with patients in the near future.
Since it was not part of this work to carry out an extensive multi-feature or multi-classifier comparison in terms of a benchmark, future work could address or investigate different feature and classification method combinations of ensemble approaches more comprehensively. Such an analysis could unveil the full potential of ensemble classifier approaches in asynchronous EEG classification and extend the findings of this work.

%\section*{Declarations}

%%%%%%%%%%%%%%%%%%%%%%%%%%%%%%%%%%%%%%%%%%%%%%
%%                                          %%
%% Backmatter begins here                   %%
%%                                          %%
%%%%%%%%%%%%%%%%%%%%%%%%%%%%%%%%%%%%%%%%%%%%%%

\begin{backmatter}

\section*{Acknowledgements}%% if any
The authors would like to thank Su Kyoung Kim for the support and feedback on the statistical analyses, and Sabisan Santhakumaran for inputs on the readability and comprehensibility of the document. Furthermore, we acknowledge support by the Open Access Publication fund of the University of Duisburg-Essen. The authors declare that they have no affiliations with or involvement in any organization or entity with any financial interest in the subject matter or materials discussed in this manuscript.

\section*{Funding}%% if any
This work was funded by the German Federal Ministry of Research, Technology and Space within the project NEARBY (Grant number: 01IS23073).

\section*{Data availability}%% if any
The datasets generated and/or analyzed during the current study are available in the following Zenodo repository: \url{https://doi.org/10.5281/zenodo.17940098}. 

\section*{Author contributions}
\textit{Conceptualization}: N.K. and E.A.K. conceptualized the study. \\ \noindent
\textit{Data curation:} N.K. performt the data curation. \\ \noindent
\textit{Formal analysis:} N.K. and K.C. were involved in the formal analysis. \\ \noindent
\textit{Funding acquisition:} E.A.K managed the founding acquisition. \\ \noindent
\textit{Investigation:} N.K. conducted the experiment and recorded the data. \\ \noindent
\textit{Methodology:} All authors discussed and designed the applied methods of this work. \\ \noindent
\textit{Project administration, Resources and Supervision:} E.A.K handled the project administration, availability of resources (lab equipment etc.) and supervised the study. \\ \noindent
\textit{Software and Visualization:} N.K. wrote the software for data analysis, evaluations and created the visualizations of the work. \\ \noindent
\textit{Validation:} N.K. and E.A.K verified the results of the study. \\ \noindent
\textit{Writing:} All authors contributed in writing and reviewing the manuscript. 

\bibliographystyle{bmc-mathphys} % Style BST file (bmc-mathphys, vancouver, spbasic).
\bibliography{bmc_article}      % Bibliography file (usually '*.bib' )

@article{kueper_avoidance_2024,
	title = {Avoidance of specific calibration sessions in motor intention recognition for exoskeleton-supported rehabilitation through transfer learning on {EEG} data},
	volume = {14},
	copyright = {2024 The Author(s)},
	issn = {2045-2322},
	url = {https://www.nature.com/articles/s41598-024-65910-8},
	doi = {10.1038/s41598-024-65910-8},
	language = {en},
	number = {1},
	urldate = {2024-07-22},
	journal = {Scientific Reports},
	author = {Kueper, Niklas and Kim, Su Kyoung and Kirchner, Elsa Andrea},
	month = jul,
	year = {2024},
	pages = {16690},
}

@article{pfurtscheller1999event,
title = {Event-related EEG/MEG synchronization and desynchronization: basic principles},
journal = {Clinical Neurophysiology},
volume = {110},
number = {11},
pages = {1842-1857},
year = {1999},
issn = {1388-2457},
doi = {https://doi.org/10.1016/S1388-2457(99)00141-8},
url = {https://www.sciencedirect.com/science/article/pii/S1388245799001418},
author = {G. Pfurtscheller and F.H. {Lopes da Silva}}
}

@article{rivet_xdawn_2009,
	title = {{xDAWN} {Algorithm} to {Enhance} {Evoked} {Potentials}: {Application} to {Brain}–{Computer} {Interface}},
	volume = {56},
	issn = {1558-2531},
	shorttitle = {{xDAWN} {Algorithm} to {Enhance} {Evoked} {Potentials}},
	url = {https://ieeexplore.ieee.org/abstract/document/4760273},
	doi = {10.1109/TBME.2009.2012869},
	number = {8},
	urldate = {2024-07-23},
	journal = {IEEE Transactions on Biomedical Engineering},
	author = {Rivet*, Bertrand and Souloumiac, Antoine and Attina, Virginie and Gibert, Guillaume},
	month = aug,
	year = {2009},
	pages = {2035--2043},
}

@article{lawhern_eegnet_2018,
	title = {{EEGNet}: a compact convolutional neural network for {EEG}-based brain–computer interfaces},
	volume = {15},
	issn = {1741-2560, 1741-2552},
	shorttitle = {{EEGNet}},
	url = {https://iopscience.iop.org/article/10.1088/1741-2552/aace8c},
	doi = {10.1088/1741-2552/aace8c},
	number = {5},
	urldate = {2025-09-26},
	journal = {Journal of Neural Engineering},
	author = {Lawhern, Vernon J and Solon, Amelia J and Waytowich, Nicholas R and Gordon, Stephen M and Hung, Chou P and Lance, Brent J},
	month = oct,
	year = {2018},
	pages = {056013},
}

@article{kirchner_multimodal_2014,
	title = {Multimodal {Movement} {Prediction} - {Towards} an {Individual} {Assistance} of {Patients}},
	volume = {9},
	issn = {1932-6203},
	url = {https://journals.plos.org/plosone/article?id=10.1371/journal.pone.0085060},
	doi = {10.1371/journal.pone.0085060},
	language = {en},
	number = {1},
	urldate = {2024-05-16},
	journal = {PLOS ONE},
	author = {Kirchner, Elsa Andrea and Tabie, Marc and Seeland, Anett},
	month = aug,
	year = {2014},
	pages = {e85060},
}

@article{feigin_world_2025,
	title = {World {Stroke} {Organization}: {Global} {Stroke} {Fact} {Sheet} 2025},
	volume = {20},
	issn = {1747-4930, 1747-4949},
	shorttitle = {World {Stroke} {Organization}},
	url = {https://journals.sagepub.com/doi/10.1177/17474930241308142},
	doi = {10.1177/17474930241308142},
	language = {en},
	number = {2},
	urldate = {2025-10-24},
	journal = {International Journal of Stroke},
	author = {Feigin, Valery L and Brainin, Michael and Norrving, Bo and Martins, Sheila O and Pandian, Jeyaraj and Lindsay, Patrice and F Grupper, Maria and Rautalin, Ilari},
	month = feb,
	year = {2025},
	pages = {132--144},
}

@article{ju_causes_2022,
	title = {Causes and {Trends} of {Disabilities} in {Community}-{Dwelling} {Stroke} {Survivors}: {A} {Population}-{Based} {Study}},
	volume = {15},
	issn = {1976-8753},
	shorttitle = {Causes and {Trends} of {Disabilities} in {Community}-{Dwelling} {Stroke} {Survivors}},
	url = {https://www.ncbi.nlm.nih.gov/pmc/articles/PMC9833459/},
	doi = {10.12786/bn.2022.15.e5},
	number = {1},
	urldate = {2025-07-07},
	journal = {Brain \& NeuroRehabilitation},
	author = {Ju, Yeon Woo and Lee, Jung Soo and Choi, Young-Ah and Kim, Yeo Hyung},
	month = mar,
	year = {2022},
	pmid = {36743839},
	pmcid = {PMC9833459},
	pages = {e5},
}

@article{krebs1998robot,
  title={Robot-aided neurorehabilitation},
  doi={10.1109/86.662623}, 
  url={https://ieeexplore.ieee.org/abstract/document/662623/references#references}, 
  author={Krebs, H Igo and Hogan, Neville and Aisen, Mindy L and Volpe, Bruce T},
  journal={IEEE transactions on rehabilitation engineering},
  volume={6},
  number={1},
  pages={75--87},
  year={1998},
  publisher={IEEE}
}

@article{mehrholz2018electromechanical,
author = {Mehrholz, Jan and Pohl, Marcus and Platz, Thomas and Kugler, Joachim and Elsner, Bernhard},
title = {Electromechanical and robot‐assisted arm training for improving activities of daily living, arm function, and arm muscle strength after stroke},
journal = {Cochrane Database of Systematic Reviews},
number = {9},
year = {2018},
publisher = {John Wiley & Sons, Ltd},
ISSN = {1465-1858},
DOI = {10.1002/14651858.CD006876.pub5},
URL = {https://doi.org//10.1002/14651858.CD006876.pub5},
}

@article{fazekas2019future,
author = {Gabor Fazekas and Ibolya Tavaszi},
title = {The future role of robots in neuro-rehabilitation},
journal = {Expert Review of Neurotherapeutics},
volume = {19},
number = {6},
pages = {471--473},
year = {2019},
publisher = {Taylor \& Francis},
doi = {10.1080/14737175.2019.1617700},
note ={PMID: 31090484},
URL = {https://doi.org/10.1080/14737175.2019.1617700},
}

@article{elsa_andrea_kirchner_towards_2022,
	title = {Towards {Bidirectional} and {Coadaptive} {Robotic} {Exoskeletons} for {Neuromotor} {Rehabilitation} and {Assisted} {Daily} {Living}: a {Review}},
	doi = {10.1007/s43154-022-00076-7},
	journal = {Current Robotics Reports},
	author = {Kirchner, Elsa Andrea and Bütefür, Judith},
	month = apr,
	year = {2022},
	note = {MAG ID: 4224283032},
    url={https://link.springer.com/article/10.1007/s43154-022-00076-7},
}

@inproceedings{noda2012brain,
  title={Brain-controlled exoskeleton robot for BMI rehabilitation},
  author={Noda, Tomoyuki and Sugimoto, Norikazu and Furukawa, Junichiro and Sato, Masa-aki and Hyon, SangHo and Morimoto, Jun},
  booktitle={2012 12th ieee-ras international conference on humanoid robots (humanoids 2012)},
  pages={21--27},
  year={2012},
  organization={IEEE}, 
  doi={10.1109/HUMANOIDS.2012.6651494}, 
  url={https://ieeexplore.ieee.org/abstract/document/6651494},
}

@article{hortal2015using,
  title={Using a brain-machine interface to control a hybrid upper limb exoskeleton during rehabilitation of patients with neurological conditions},
  author={Hortal, Enrique and Planelles, Daniel and Resquin, Francisco and Climent, Jos{\'e} M and Azor{\'\i}n, Jos{\'e} M and Pons, Jos{\'e} L},
  journal={Journal of neuroengineering and rehabilitation},
  volume={12},
  number={1},
  pages={92},
  year={2015},
  publisher={Springer}, 
  doi={https://doi.org/10.1186/s12984-015-0082-9}, 
  url={https://link.springer.com/article/10.1186/s12984-015-0082-9},
}

@article{singh2021evidence,
  title={Evidence of neuroplasticity with robotic hand exoskeleton for post-stroke rehabilitation: a randomized controlled trial},
  author={Singh, Neha and Saini, Megha and Kumar, Nand and Srivastava, MV Padma and Mehndiratta, Amit},
  journal={Journal of neuroengineering and rehabilitation},
  volume={18},
  number={1},
  pages={76},
  year={2021},
  publisher={Springer}, 
  doi={https://doi.org/10.1186/s12984-021-00867-7},
  url={https://link.springer.com/article/10.1186/s12984-021-00867-7},
}

@article{kumar2019modular,
  title={Modular design and decentralized control of the recupera exoskeleton for stroke rehabilitation},
  author={Kumar, Shivesh and W{\"o}hrle, Hendrik and Trampler, Mathias and Simnofske, Marc and Peters, Heiner and Mallwitz, Martin and Kirchner, Elsa Andrea and Kirchner, Frank},
  journal={Applied Sciences},
  volume={9},
  number={4},
  pages={626},
  year={2019},
  publisher={MDPI}, 
  doi={10.3390/app9040626}, 
  url={https://www.mdpi.com/2076-3417/9/4/626},
  issn={2076-3417},
}

@inproceedings{seeland_online_2013,
	title = {Online movement prediction in a robotic application scenario},
	doi = {10.1109/NER.2013.6695866},
	urldate = {2024-05-16},
	booktitle = {2013 6th {International} {IEEE}/{EMBS} {Conference} on {Neural} {Engineering} ({NER})},
	author = {Seeland, Anett and Woehrle, Hendrik and Straube, Sirko and Kirchner, Elsa Andrea},
	month = nov,
	year = {2013},
	note = {ISSN: 1948-3554},
	pages = {41--44},
}

@article{vahdat_single_2019,
	title = {A {Single} {Session} of {Robot}-{Controlled} {Proprioceptive} {Training} {Modulates} {Functional} {Connectivity} of {Sensory} {Motor} {Networks} and {Improves} {Reaching} {Accuracy} in {Chronic} {Stroke}},
	volume = {33},
	issn = {1545-9683, 1552-6844},
	url = {https://journals.sagepub.com/doi/10.1177/1545968318818902},
	doi = {10.1177/1545968318818902},
	language = {en},
	number = {1},
	urldate = {2025-11-13},
	journal = {Neurorehabilitation and Neural Repair},
	author = {Vahdat, Shahabeddin and Darainy, Mohammed and Thiel, Alexander and Ostry, David J.},
	month = jan,
	year = {2019},
	pages = {70--81},
}

@article{liu_eeg-based_2018,
	title = {{EEG}-{Based} {Lower}-{Limb} {Movement} {Onset} {Decoding}: {Continuous} {Classification} and {Asynchronous} {Detection}},
	volume = {26},
	issn = {1558-0210},
	shorttitle = {{EEG}-{Based} {Lower}-{Limb} {Movement} {Onset} {Decoding}},
	url = {https://ieeexplore.ieee.org/abstract/document/8410007},
	doi = {10.1109/TNSRE.2018.2855053},
	number = {8},
	urldate = {2025-07-14},
	journal = {IEEE Transactions on Neural Systems and Rehabilitation Engineering},
	author = {Liu, Dong and Chen, Weihai and Lee, Kyuhwa and Chavarriaga, Ricardo and Iwane, Fumiaki and Bouri, Mohamed and Pei, Zhongcai and Millán, José del R.},
	month = aug,
	year = {2018},
	pages = {1626--1635},
}

@article{bai_prediction_2011,
	title = {Prediction of human voluntary movement before it occurs},
	volume = {122},
	issn = {1388-2457},
	url = {https://www.ncbi.nlm.nih.gov/pmc/articles/PMC5558611/},
	doi = {10.1016/j.clinph.2010.07.010},
	number = {2},
	urldate = {2025-02-12},
	journal = {Clinical neurophysiology : official journal of the International Federation of Clinical Neurophysiology},
	author = {Bai, Ou and Rathi, Varun and Lin, Peter and Huang, Dandan and Battapady, Harsha and Fei, Ding-Yu and Schneider, Logan and Houdayer, Elise and Chen, Xuedong and Hallett, Mark},
	month = feb,
	year = {2011},
	pmid = {20675187},
	pmcid = {PMC5558611},
	pages = {364--372},
}

@article{sburlea_continuous_2015,
	title = {Continuous detection of the self-initiated walking pre-movement state from {EEG} correlates without session-to-session recalibration},
	volume = {12},
	issn = {1741-2560, 1741-2552},
	url = {https://iopscience.iop.org/article/10.1088/1741-2560/12/3/036007},
	doi = {10.1088/1741-2560/12/3/036007},
	number = {3},
	urldate = {2025-10-03},
	journal = {Journal of Neural Engineering},
	author = {Sburlea, Andreea Ioana and Montesano, Luis and Minguez, Javier},
	month = jun,
	year = {2015},
	pages = {036007},
}

@inproceedings{kuo_classification_2011,
	title = {Classification of intended motor movement using surface {EEG} ensemble empirical mode decomposition},
	doi = {10.1109/IEMBS.2011.6091550},
	urldate = {2025-01-02},
	booktitle = {2011 {Annual} {International} {Conference} of the {IEEE} {Engineering} in {Medicine} and {Biology} {Society}},
	author = {Kuo, Ching-Chang and Lin, William S. and Dressel, Chelsea A. and Chiu, Alan W. L.},
	month = aug,
	year = {2011},
	note = {ISSN: 1558-4615},
	pages = {6281--6284},
}

@article{lopez-larraz_continuous_2014,
	title = {Continuous decoding of movement intention of upper limb self-initiated analytic movements from pre-movement {EEG} correlates},
	volume = {11},
	issn = {1743-0003},
	url = {https://doi.org/10.1186/1743-0003-11-153},
	doi = {10.1186/1743-0003-11-153},
	language = {en},
	number = {1},
	urldate = {2024-08-17},
	journal = {Journal of NeuroEngineering and Rehabilitation},
	author = {L{\'o}pez-Larraz, Eduardo and Montesano, Luis and Gil-Agudo, {\'A}ngel and Minguez, Javier},
	month = nov,
	year = {2014},
	pages = {153},
}

@article{lew_detection_2012,
	title = {Detection of self-paced reaching movement intention from {EEG} signals},
	volume = {5},
	issn = {1662-6443},
	url = {https://www.frontiersin.org/journals/neuroengineering/articles/10.3389/fneng.2012.00013/full},
	doi = {10.3389/fneng.2012.00013},
	language = {English},
	urldate = {2025-07-10},
	journal = {Frontiers in Neuroengineering},
	author = {Lew, Eileen and Chavarriaga, Ricardo and Silvoni, Stefano and Millán, José del R.},
	month = jul,
	year = {2012},
}

@article{lana_detection_2015,
	title = {Detection of movement intention using {EEG} in a human-robot interaction environment},
	volume = {31},
	issn = {2446-4732, 2446-4740},
	url = {https://www.scielo.br/j/reng/a/j6nWVKbxgRjNYkF87SbXVBD/?lang=en},
	doi = {10.1590/2446-4740.0777},
	language = {en},
	urldate = {2024-03-08},
	journal = {Research on Biomedical Engineering},
	author = {Lana, Ernesto Pablo and Adorno, Bruno Vilhena and Tierra-Criollo, Carlos Julio},
	month = nov,
	year = {2015},
	pages = {285--294},
}

@article{ceradini_effect_2025,
	title = {The {Effect} of {User} {Learning} for {Online} {EEG} {Decoding} of {Upper}-{Limb} {Movement} {Intention}},
	volume = {7},
	issn = {2576-3202},
	url = {https://ieeexplore.ieee.org/abstract/document/10869338},
	doi = {10.1109/TMRB.2025.3537663},
	number = {2},
	urldate = {2025-10-03},
	journal = {IEEE Transactions on Medical Robotics and Bionics},
	author = {Ceradini, Matteo and Tortora, Stefano and Micera, Silvestro and Tonin, Luca},
	month = may,
	year = {2025},
	pages = {633--641},
}

@article{sarasola-sanz_hybrid_2024,
	title = {A hybrid brain-muscle-machine interface for stroke rehabilitation: {Usability} and functionality validation in a 2-week intensive intervention},
	volume = {12},
	issn = {2296-4185},
	shorttitle = {A hybrid brain-muscle-machine interface for stroke rehabilitation},
	url = {https://www.frontiersin.org/journals/bioengineering-and-biotechnology/articles/10.3389/fbioe.2024.1330330/full},
	doi = {10.3389/fbioe.2024.1330330},
	language = {English},
	urldate = {2025-01-24},
	journal = {Frontiers in Bioengineering and Biotechnology},
	author = {Sarasola-Sanz, Andrea and Ray, Andreas M. and Insausti-Delgado, Ainhoa and Irastorza-Landa, Nerea and Mahmoud, Wala Jaser and Brötz, Doris and Bibián-Nogueras, Carlos and Helmhold, Florian and Zrenner, Christoph and Ziemann, Ulf and L{\'o}pez-Larraz, Eduardo and Ramos-Murguialday, Ander},
	month = apr,
	year = {2024},
}

@article{gratton_pre-_1988,
	title = {Pre- and poststimulus activation of response channels: {A} psychophysiological analysis},
	volume = {14},
	issn = {1939-1277},
	shorttitle = {Pre- and poststimulus activation of response channels},
	doi = {10.1037/0096-1523.14.3.331},
	number = {3},
	journal = {Journal of Experimental Psychology: Human Perception and Performance},
	author = {Gratton, Gabriele and Coles, Michael G. H. and Sirevaag, Erik J. and Eriksen, Charles W. and Donchin, Emanuel},
	year = {1988},
	pages = {331--344},
}

@article{de_jong_use_1988,
	title = {Use of partial stimulus information in response processing.},
	volume = {14},
	issn = {1939-1277, 0096-1523},
	url = {https://doi.apa.org/doi/10.1037/0096-1523.14.4.682},
	doi = {10.1037/0096-1523.14.4.682},
	language = {en},
	number = {4},
	urldate = {2025-11-13},
	journal = {Journal of Experimental Psychology: Human Perception and Performance},
	author = {De Jong, Ritske and Wierda, Marcel and Mulder, Gijsbertus and Mulder, Lambertus J.},
	year = {1988},
	pages = {682--692},
}

@article{kornhuber_hirnpotentialanderungen_1965,
	title = {Hirnpotentialänderungen bei {Willkürbewegungen} und passiven {Bewegungen} des {Menschen}: {Bereitschaftspotential} und reafferente {Potentiale}},
	volume = {284},
	issn = {1432-2013},
	shorttitle = {Hirnpotentialänderungen bei {Willkürbewegungen} und passiven {Bewegungen} des {Menschen}},
	url = {https://doi.org/10.1007/BF00412364},
	doi = {10.1007/BF00412364},
	language = {de},
	number = {1},
	urldate = {2025-10-30},
	journal = {Pflüger's Archiv für die gesamte Physiologie des Menschen und der Tiere},
	author = {Kornhuber, Hans H. and Deecke, Lüder},
	month = mar,
	year = {1965},
	pages = {1--17},
}

@article{shibasaki_what_2006,
	title = {What is the {Bereitschaftspotential}?},
	volume = {117},
	issn = {1388-2457},
	url = {https://www.sciencedirect.com/science/article/pii/S138824570600229X},
	doi = {10.1016/j.clinph.2006.04.025},
	number = {11},
	urldate = {2025-10-30},
	journal = {Clinical Neurophysiology},
	author = {Shibasaki, Hiroshi and Hallett, Mark},
	month = nov,
	year = {2006},
	pages = {2341--2356},
}

@article{pfurtscheller_event-related_1999,
	title = {Event-related {EEG}/{MEG} synchronization and desynchronization: basic principles},
	volume = {110},
	issn = {1388-2457},
	shorttitle = {Event-related {EEG}/{MEG} synchronization and desynchronization},
	url = {https://www.sciencedirect.com/science/article/pii/S1388245799001418},
	doi = {10.1016/S1388-2457(99)00141-8},
	number = {11},
	urldate = {2024-10-11},
	journal = {Clinical Neurophysiology},
	author = {Pfurtscheller, G. and Lopes da Silva, F. H.},
	month = nov,
	year = {1999},
	pages = {1842--1857},
}

@inproceedings{seeland_spatio-temporal_2015,
	title = {Spatio-temporal {Comparison} between {ERD}/{ERS} and {MRCP}-based {Movement} {Prediction}},
	volume = {2},
	isbn = {9789897580697},
	url = {https://www.scitepress.org/PublishedPapers/2015/52140},
	doi = {10.5220/0005214002190226},
    booktitle={International Conference on Bio-inspired Systems and Signal Processing},
	language = {en},
	urldate = {2024-10-11},
	publisher = {SCITEPRESS},
	author = {Seeland, Anett and Manca, Laura and Kirchner, Frank and Kirchner, Elsa Andrea},
	month = jan,
	year = {2015},
	pages = {219--226},
}

@article{lopez-larraz_uncovering_2025,
	title = {Uncovering attempted movements of the paralyzed upper limb after stroke through {EEG} and {EMG}},
	volume = {22},
	issn = {1743-0003},
	url = {https://doi.org/10.1186/s12984-025-01687-9},
	doi = {10.1186/s12984-025-01687-9},
	language = {en},
	number = {1},
	urldate = {2025-11-07},
	journal = {Journal of NeuroEngineering and Rehabilitation},
	author = {L{\'o}pez-Larraz, Eduardo and Sarasola-Sanz, Andrea and Birbaumer, Niels and Ramos-Murguialday, Ander},
	month = oct,
	year = {2025},
	pages = {221},
}

@article{shakeel_review_2015,
	title = {A {Review} of {Techniques} for {Detection} of {Movement} {Intention} {Using} {Movement}-{Related} {Cortical} {Potentials}},
	volume = {2015},
	copyright = {http://creativecommons.org/licenses/by/4.0/},
	issn = {1748-670X, 1748-6718},
	url = {http://www.hindawi.com/journals/cmmm/2015/346217/},
	doi = {10.1155/2015/346217},
	language = {en},
	urldate = {2025-11-13},
	journal = {Computational and Mathematical Methods in Medicine},
	author = {Shakeel, Aqsa and Navid, Muhammad Samran and Anwar, Muhammad Nabeel and Mazhar, Suleman and Jochumsen, Mads and Niazi, Imran Khan},
	year = {2015},
	pages = {1--13},
}

@inproceedings{ang_clinical_2009,
	title = {A clinical study of motor imagery-based brain-computer interface for upper limb robotic rehabilitation},
	url = {https://ieeexplore.ieee.org/abstract/document/5335381},
	doi = {10.1109/IEMBS.2009.5335381},
	urldate = {2025-11-13},
	booktitle = {2009 {Annual} {International} {Conference} of the {IEEE} {Engineering} in {Medicine} and {Biology} {Society}},
	author = {Ang, Kai Keng and Guan, Cuntai and Chua, Karen Sui Geok and Ang, Beng Ti and Kuah, Christopher and Wang, Chuanchu and Phua, Kok Soon and Chin, Zheng Yang and Zhang, Haihong},
	month = sep,
	year = {2009},
	note = {ISSN: 1558-4615},
	pages = {5981--5984},
}

@article{lopez-larraz_brain-machine_2018,
	title = {Brain-machine interfaces for rehabilitation in stroke: {A} review},
	volume = {43},
	issn = {10538135, 18786448},
	shorttitle = {Brain-machine interfaces for rehabilitation in stroke},
	url = {https://journals.sagepub.com/doi/full/10.3233/NRE-172394},
	doi = {10.3233/NRE-172394},
	number = {1},
	urldate = {2025-11-13},
	journal = {NeuroRehabilitation},
	author = {L{\'o}pez-Larraz, E. and Sarasola-Sanz, A. and Irastorza-Landa, N. and Birbaumer, N. and Ramos-Murguialday, A.},
	editor = {Harvey, Richard L.},
	month = jul,
	year = {2018},
	pages = {77--97},
}

@article{yang_eeg_2022,
	title = {{EEG}- and {EMG}-{Driven} {Poststroke} {Rehabilitation}: {A} {Review}},
	volume = {22},
	issn = {1558-1748},
	shorttitle = {{EEG}- and {EMG}-{Driven} {Poststroke} {Rehabilitation}},
	url = {https://ieeexplore.ieee.org/abstract/document/9950491?casa\_token=-Q\_zLVjLhtkAAAAA:MrSXHlWa5Q\_p1If6F1nMDGXgGIJxQn9FCLwyDID0O9p2QdgWZVO7UlLfxDN265IRU6H6nee3YgE},
	doi = {10.1109/JSEN.2022.3220930},
	number = {24},
	urldate = {2024-08-17},
	journal = {IEEE Sensors Journal},
	author = {Yang, Haiyang and Wan, Jiacheng and Jin, Ying and Yu, Xixia and Fang, Yinfeng},
	month = dec,
	year = {2022},
	pages = {23649--23660},
}

@article{lotte_review_2018,
	title = {A review of classification algorithms for {EEG}-based brain–computer interfaces: a 10 year update},
	volume = {15},
	issn = {1741-2560, 1741-2552},
	shorttitle = {A review of classification algorithms for {EEG}-based brain–computer interfaces},
	url = {https://iopscience.iop.org/article/10.1088/1741-2552/aab2f2},
	doi = {10.1088/1741-2552/aab2f2},
	number = {3},
	urldate = {2025-10-03},
	journal = {Journal of Neural Engineering},
	author = {Lotte, F and Bougrain, L and Cichocki, A and Clerc, M and Congedo, M and Rakotomamonjy, A and Yger, F},
	month = jun,
	year = {2018},
	pages = {031005},
}

@article{lotte_review_2007,
	title = {A review of classification algorithms for {EEG}-based brain–computer interfaces},
	volume = {4},
	issn = {1741-2560, 1741-2552},
	url = {https://iopscience.iop.org/article/10.1088/1741-2560/4/2/R01},
	doi = {10.1088/1741-2560/4/2/R01},
	number = {2},
	urldate = {2025-10-03},
	journal = {Journal of Neural Engineering},
	author = {Lotte, F and Congedo, M and Lécuyer, A and Lamarche, F and Arnaldi, B},
	month = jun,
	year = {2007},
	pages = {R1--R13},
}

@article{liu_hybrid_2024,
	title = {A hybrid method for asynchronous detection of motor imagery electroencephalogram fusing alpha rhythm and movement-related cortical potential},
	volume = {237},
	issn = {0263-2241},
	url = {https://www.sciencedirect.com/science/article/pii/S0263224124010522},
	doi = {10.1016/j.measurement.2024.115167},
	urldate = {2025-10-24},
	journal = {Measurement},
	author = {Liu, Xiaolin and Sun, Ying and Wang, Shuai and Yan, Jun and Jiang, Ziyu and Zheng, Dezhi},
	month = sep,
	year = {2024},
	pages = {115167},
}

@article{luo_motor_2020,
	title = {Motor imagery {EEG} classification based on ensemble support vector learning},
	volume = {193},
	issn = {0169-2607},
	url = {https://www.sciencedirect.com/science/article/pii/S0169260719317511},
	doi = {10.1016/j.cmpb.2020.105464},
	urldate = {2025-10-03},
	journal = {Computer Methods and Programs in Biomedicine},
	author = {Luo, Jing and Gao, Xing and Zhu, Xiaobei and Wang, Bin and Lu, Na and Wang, Jie},
	month = sep,
	year = {2020},
	pages = {105464},
}

@article{chen_toward_2024,
	title = {Toward reliable signals decoding for electroencephalogram: {A} benchmark study to {EEGNeX}},
	volume = {87},
	issn = {1746-8094},
	shorttitle = {Toward reliable signals decoding for electroencephalogram},
	url = {https://www.sciencedirect.com/science/article/pii/S1746809423009084},
	doi = {10.1016/j.bspc.2023.105475},
	urldate = {2025-11-14},
	journal = {Biomedical Signal Processing and Control},
	author = {Chen, Xia and Teng, Xiangbin and Chen, Han and Pan, Yafeng and Geyer, Philipp},
	month = jan,
	year = {2024},
	pages = {105475},
}

@inproceedings{arpaia_sinc-eegnet_2023,
	title = {Sinc-{EEGNet} for {Improving} {Performance} {While} {Reducing} {Calibration} of a {Motor} {Imagery}-{Based} {BCI}},
	url = {https://ieeexplore.ieee.org/document/10405701},
	doi = {10.1109/MetroXRAINE58569.2023.10405701},
	urldate = {2024-11-21},
	booktitle = {2023 {IEEE} {International} {Conference} on {Metrology} for {eXtended} {Reality}, {Artificial} {Intelligence} and {Neural} {Engineering} ({MetroXRAINE})},
	author = {Arpaia, Pasquale and Bertone, Elisa and Esposito, Antonio and Natalizio, Angela and Parvis, Marco and Giulia Pedrocchi, Alessandra Laura and Pollastro, Andrea},
	month = oct,
	year = {2023},
	pages = {1063--1068},
}

@article{schirrmeister_deep_2017,
	title = {Deep learning with convolutional neural networks for {EEG} decoding and visualization},
	volume = {38},
	copyright = {http://creativecommons.org/licenses/by/4.0/},
	issn = {1065-9471, 1097-0193},
	url = {https://onlinelibrary.wiley.com/doi/10.1002/hbm.23730},
	doi = {10.1002/hbm.23730},
	language = {en},
	number = {11},
	urldate = {2025-11-14},
	journal = {Human Brain Mapping},
	author = {Schirrmeister, Robin Tibor and Springenberg, Jost Tobias and Fiederer, Lukas Dominique Josef and Glasstetter, Martin and Eggensperger, Katharina and Tangermann, Michael and Hutter, Frank and Burgard, Wolfram and Ball, Tonio},
	month = nov,
	year = {2017},
	pages = {5391--5420},
}

@article{zeynali_classification_2023,
	title = {Classification of {EEG} signals using Transformer based deep learning and ensemble models},
	volume = {86},
	issn = {1746-8094},
	url = {https://www.sciencedirect.com/science/article/pii/S1746809423005633},
	doi = {10.1016/j.bspc.2023.105130},
	pages = {105130},
	journaltitle = {Biomedical Signal Processing and Control},
	shortjournal = {Biomedical Signal Processing and Control},
	journal = {Biomedical Signal Processing and Control},
	author = {Zeynali, Mahsa and Seyedarabi, Hadi and Afrouzian, Reza},
	urldate = {2025-10-03},
	date = {2023-09-01},
	year = {2023},
}

@article{pfurtscheller_graz-bci_2003,
	title = {Graz-{BCI}: state of the art and clinical applications},
	volume = {11},
	issn = {1558-0210},
	shorttitle = {Graz-{BCI}},
	url = {https://ieeexplore.ieee.org/abstract/document/1214714},
	doi = {10.1109/TNSRE.2003.814454},
	number = {2},
	urldate = {2025-11-08},
	journal = {IEEE Transactions on Neural Systems and Rehabilitation Engineering},
	author = {Pfurtscheller, G. and Neuper, C. and Muller, G.R. and Obermaier, B. and Krausz, G. and Schlogl, A. and Scherer, R. and Graimann, B. and Keinrath, C. and Skliris, D. and Wortz, M. and Supp, G. and Schrank, C.},
	month = jun,
	year = {2003},
	pages = {1--4},
}

@article{song_eeg-based_2022,
	title = {An {EEG}-based asynchronous {MI}-{BCI} system to reduce false positives with a small number of channels for neurorehabilitation: {A} pilot study},
	volume = {16},
	issn = {1662-5218},
	shorttitle = {An {EEG}-based asynchronous {MI}-{BCI} system to reduce false positives with a small number of channels for neurorehabilitation},
	url = {https://www.frontiersin.org/journals/neurorobotics/articles/10.3389/fnbot.2022.971547/full},
	doi = {10.3389/fnbot.2022.971547},
	language = {English},
	urldate = {2025-10-24},
	journal = {Frontiers in Neurorobotics},
	author = {Song, Minsu and Jeong, Hojun and Kim, Jongbum and Jang, Sung-Ho and Kim, Jonghyun},
	month = sep,
	year = {2022},
}

@article{sun_experimental_2007,
	title = {An experimental evaluation of ensemble methods for {EEG} signal classification},
	volume = {28},
	issn = {0167-8655},
	url = {https://www.sciencedirect.com/science/article/pii/S0167865507002085},
	doi = {10.1016/j.patrec.2007.06.018},
	number = {15},
	urldate = {2024-10-11},
	journal = {Pattern Recognition Letters},
	author = {Sun, Shiliang and Zhang, Changshui and Zhang, Dan},
	month = nov,
	year = {2007},
	pages = {2157--2163},
}

@incollection{soria-frisch_critical_2013,
	address = {Berlin, Heidelberg},
	title = {A {Critical} {Review} on the {Usage} of {Ensembles} for {BCI}},
	isbn = {9783642297465},
	url = {https://doi.org/10.1007/978-3-642-29746-5\_3},
	language = {en},
	urldate = {2025-10-03},
	booktitle = {Towards {Practical} {Brain}-{Computer} {Interfaces}: {Bridging} the {Gap} from {Research} to {Real}-{World} {Applications}},
	publisher = {Springer},
	author = {Soria-Frisch, Aureli},
	editor = {Allison, Brendan Z. and Dunne, Stephen and Leeb, Robert and Del R. Millán, José and Nijholt, Anton},
	year = {2013},
	doi = {10.1007/978-3-642-29746-5\_3},
	pages = {41--65},
}

@article{alsuradi_ensemble_2022,
	title = {An ensemble deep learning approach to evaluate haptic delay from a single trial {EEG} data},
	volume = {9},
	issn = {2296-9144},
	url = {https://www.frontiersin.org/journals/robotics-and-ai/articles/10.3389/frobt.2022.1013043/full},
	doi = {10.3389/frobt.2022.1013043},
	language = {English},
	urldate = {2025-10-03},
	journal = {Frontiers in Robotics and AI},
	author = {Alsuradi, Haneen and Eid, Mohamad},
	month = sep,
	year = {2022},
}

@article{dhara_fuzzy_2024,
	title = {A {Fuzzy} {Ensemble}-{Based} {Deep} learning {Model} for {EEG}-{Based} {Emotion} {Recognition}},
	volume = {16},
	issn = {1866-9964},
	url = {https://doi.org/10.1007/s12559-023-10171-2},
	doi = {10.1007/s12559-023-10171-2},
	language = {en},
	number = {3},
	urldate = {2024-10-11},
	journal = {Cognitive Computation},
	author = {Dhara, Trishita and Singh, Pawan Kumar and Mahmud, Mufti},
	month = may,
	year = {2024},
	pages = {1364--1378},
}

@article{koley_ensemble_2012,
	title = {An ensemble system for automatic sleep stage classification using single channel {EEG} signal},
	volume = {42},
	issn = {0010-4825},
	url = {https://www.sciencedirect.com/science/article/pii/S0010482512001588},
	doi = {10.1016/j.compbiomed.2012.09.012},
	number = {12},
	urldate = {2024-10-11},
	journal = {Computers in Biology and Medicine},
	author = {Koley, B. and Dey, D.},
	month = dec,
	year = {2012},
	pages = {1186--1195},
}

@article{alsuradi_ensemble_2023,
	title = {An ensemble deep-learning approach for single-trial {EEG} classification of vibration intensity},
	volume = {20},
	issn = {1741-2560, 1741-2552},
	url = {https://iopscience.iop.org/article/10.1088/1741-2552/acfbf9},
	doi = {10.1088/1741-2552/acfbf9},
	number = {5},
	urldate = {2025-01-06},
	journal = {Journal of Neural Engineering},
	author = {Alsuradi, Haneen and Park, Wanjoo and Eid, Mohamad},
	month = oct,
	year = {2023},
	pages = {056027},
}

@article{abbasi_eeg-based_2021,
	title = {{EEG}-{Based} {Neonatal} {Sleep} {Stage} {Classification} {Using} {Ensemble} {Learning}},
	volume = {70},
	issn = {1546-2218},
	doi = {10.32604/cmc.2022.020318},
	number = {3},
	journal = {Computers Materials \& Continua},
	author = {Abbasi, Saadullah Farooq and Jamil, Harun and Chen, Wei},
	month = oct,
	year = {2021},
	pages = {4619--4633},
    url={https://research.birmingham.ac.uk/en/publications/eeg-based-neonatal-sleep-stage-classification-using-ensemble-lear/}, 
}

@article{prabhakar_ensemble_2024,
	title = {Ensemble {Fusion} {Models} {Using} {Various} {Strategies} and {Machine} {Learning} for {EEG} {Classification}},
	volume = {11},
	copyright = {http://creativecommons.org/licenses/by/3.0/},
	issn = {2306-5354},
	url = {https://www.mdpi.com/2306-5354/11/10/986},
	doi = {10.3390/bioengineering11100986},
	language = {en},
	number = {10},
	urldate = {2024-10-11},
	journal = {Bioengineering},
	author = {Prabhakar, Sunil Kumar and Lee, Jae Jun and Won, Dong-Ok},
	month = oct,
	year = {2024},
	pages = {986},
}

@article{hosseini_random_2018,
	title = {Random ensemble learning for {EEG} classification},
	volume = {84},
	issn = {0933-3657},
	url = {https://www.sciencedirect.com/science/article/pii/S0933365717302014},
	doi = {10.1016/j.artmed.2017.12.004},
	urldate = {2024-10-11},
	journal = {Artificial Intelligence in Medicine},
	author = {Hosseini, Mohammad-Parsa and Pompili, Dario and Elisevich, Kost and Soltanian-Zadeh, Hamid},
	month = jan,
	year = {2018},
	pages = {146--158},
}

@article{zheng_ensemble_2022,
	title = {Ensemble learning method based on temporal, spatial features with multi-scale filter banks for motor imagery {EEG} classification},
	volume = {76},
	issn = {1746-8094},
	url = {https://www.sciencedirect.com/science/article/pii/S1746809422001562},
	doi = {10.1016/j.bspc.2022.103634},
	urldate = {2025-10-03},
	journal = {Biomedical Signal Processing and Control},
	author = {Zheng, Liangsheng and Feng, Wei and Ma, Yue and Lian, Pengchen and Xiao, Yang and Yi, Zhengkun and Wu, Xinyu},
	month = jul,
	year = {2022},
	pages = {103634},
}

@inproceedings{bhattacharyya_performance_2014,
	title = {Performance analysis of ensemble methods for multi-class classification of motor imagery {EEG} signal},
	url = {https://ieeexplore.ieee.org/document/6959183/},
	doi = {10.1109/CIEC.2014.6959183},
	urldate = {2025-10-03},
	booktitle = {Proceedings of {The} 2014 {International} {Conference} on {Control}, {Instrumentation}, {Energy} and {Communication} ({CIEC})},
	author = {Bhattacharyya, Saugat and Konar, Amit and Tibarewala, D. N. and Khasnobish, Anwesha and Janarthanan, R.},
	month = jan,
	year = {2014},
	pages = {712--716},
}

@article{dolzhikova_subject_independent_2022,
	title = {Subject-{Independent} {Classification} of {Motor} {Imagery} {Tasks} in {EEG} {Using} {Multisubject} {Ensemble} {CNN}},
	volume = {10},
	issn = {2169-3536},
	url = {https://ieeexplore.ieee.org/abstract/document/9846975},
	doi = {10.1109/ACCESS.2022.3195513},
	urldate = {2025-10-03},
	journal = {IEEE Access},
	author = {Dolzhikova, Irina and Abibullaev, Berdakh and Sameni, Reza and Zollanvari, Amin},
	year = {2022},
	pages = {81355--81363},
}

@article{gaur_sliding_2021,
	title = {A {Sliding} {Window} {Common} {Spatial} {Pattern} for {Enhancing} {Motor} {Imagery} {Classification} in {EEG}-{BCI}},
	volume = {70},
	issn = {1557-9662},
	url = {https://ieeexplore.ieee.org/abstract/document/9326392},
	doi = {10.1109/TIM.2021.3051996},
	urldate = {2025-10-03},
	journal = {IEEE Transactions on Instrumentation and Measurement},
	author = {Gaur, Pramod and Gupta, Harsh and Chowdhury, Anirban and McCreadie, Karl and Pachori, Ram Bilas and Wang, Hui},
	year = {2021},
	pages = {1--9},
}

@article{saideepthi_sliding_2023,
	title = {Sliding {Window} {Along} {With} {EEGNet}-{Based} {Prediction} of {EEG} {Motor} {Imagery}},
	volume = {23},
	issn = {1558-1748},
	url = {https://ieeexplore.ieee.org/abstract/document/10115246},
	doi = {10.1109/JSEN.2023.3270281},
	number = {15},
	urldate = {2025-10-03},
	journal = {IEEE Sensors Journal},
	author = {Saideepthi, Pabba and Chowdhury, Anirban and Gaur, Pramod and Pachori, Ram Bilas},
	month = aug,
	year = {2023},
	pages = {17703--17713},
}

@article{folgheraiter_measuring_2012,
	title = {Measuring the {Improvement} of the {Interaction} {Comfort} of a {Wearable} {Exoskeleton}},
	volume = {4},
	issn = {1875-4805},
	url = {https://doi.org/10.1007/s12369-012-0147-x},
	doi = {10.1007/s12369-012-0147-x},
	language = {en},
	number = {3},
	urldate = {2025-12-06},
	journal = {International Journal of Social Robotics},
	author = {Folgheraiter, Michele and Jordan, Mathias and Straube, Sirko and Seeland, Anett and Kim, Su Kyoung and Kirchner, Elsa Andrea},
	month = aug,
	year = {2012},
	pages = {285--302},
}

@inproceedings{seeland_adaptive_2017,
	title = {Adaptive multimodal biosignal control for exoskeleton supported stroke rehabilitation},
	url = {https://ieeexplore.ieee.org/abstract/document/8122987},
	doi = {10.1109/SMC.2017.8122987},
	urldate = {2025-12-06},
	booktitle = {2017 {IEEE} {International} {Conference} on {Systems}, {Man}, and {Cybernetics} ({SMC})},
	author = {Seeland, Anett and Tabie, Marc and Kim, Su Kyoung and Kirchner, Frank and Kirchner, Elsa Andrea},
	month = oct,
	year = {2017},
	pages = {2431--2436},
}

@inproceedings{kim2023asynchronous,
  title={Asynchronous classification of error-related potentials in human-robot interaction},
  author={Kim, Su Kyoung and Maurus, Michael and Trampler, Mathias and Tabie, Marc and Kirchner, Elsa Andrea},
  booktitle={International Conference on Human-Computer Interaction},
  pages={92--101},
  year={2023},
  organization={Springer},
  doi={https://doi.org/10.1007/978-3-031-35602-5\_7},
  url={https://link.springer.com/chapter/10.1007/978-3-031-35602-5\_7},

}
% for author-year bibliography (bmc-mathphys or spbasic)
% a) write to bib file (bmc-mathphys only)
% @settings{label, options="nameyear"}
% b) uncomment next line
%\nocite{label}

% or include bibliography directly:
% \begin{thebibliography}
% \bibitem{b1}
% \end{thebibliography}

%%%%%%%%%%%%%%%%%%%%%%%%%%%%%%%%%%%
%%                               %%
%% Figures                       %%
%%                               %%
%% NB: this is for captions and  %%
%% Titles. All graphics must be  %%
%% submitted separately and NOT  %%
%% included in the Tex document  %%
%%                               %%
%%%%%%%%%%%%%%%%%%%%%%%%%%%%%%%%%%%

\end{backmatter}
\end{document}